\documentclass[10pt,twocolumn,letterpaper]{article}

\usepackage{iccv}
\usepackage{times}
\usepackage{epsfig}
\usepackage{graphicx}
\usepackage{amsmath}
\usepackage{amssymb}
\usepackage{multirow}
\usepackage{booktabs}
\usepackage[
detect-weight=true, 
detect-family=true,
separate-uncertainty = true,
multi-part-units = repeat
]{siunitx}
\usepackage{amsthm}
\usepackage{pifont}%
\newcommand{\cmark}{\ding{51}}%
\usepackage[pagebackref=true,breaklinks=true,letterpaper=true,colorlinks,bookmarks=false]{hyperref}

\iccvfinalcopy 

\def\iccvPaperID{6814} 

\ificcvfinal\fi

\begin{document}
	
	\title{InstanceRefer: Cooperative Holistic Understanding for Visual Grounding on Point Clouds through Instance Multi-level Contextual Referring}
	
	\author{Zhihao Yuan$^{1, \dagger}$, Xu Yan$^{1, \dagger}$, Yinghong Liao$^{1}$,  Ruimao Zhang$^{1}$, 
		Sheng Wang$^{2}$,  Zhen Li$^{1,}$\thanks{{ Corresponding author: Zhen Li. $^\dagger$ Equal first authorship.}}~, Shuguang Cui$^{1}$ \\
		\\
		$^{1}$The Chinese University of Hong Kong (Shenzhen), Shenzhen Research Institute of Big Data \\
		$^{2}$CryoEM Center, Southern University of Science and Technology \\
		{\tt\small	\{{zhihaoyuan@link.}, xuyan1@link., {lizhen@}\}cuhk.edu.cn}}
	
	\maketitle
	\ificcvfinal\thispagestyle{empty}\fi
	
	\begin{abstract}
		Compared with the visual grounding on 2D images, the natural-language-guided 3D object localization on point clouds is more challenging.
		In this paper, we propose a new model, named \textbf{InstanceRefer}\footnote{~\url{https://github.com/CurryYuan/InstanceRefer}}, to achieve a superior 3D visual grounding through the grounding-by-matching strategy.
		In practice, our model first predicts the target category from the language descriptions using a simple language classification model.
		Then based on the category, our model sifts out a small number of instance candidates (usually less than 20) from the panoptic segmentation on point clouds.
		Thus, the non-trivial 3D visual grounding task has been effectively re-formulated as a simplified instance-matching problem, considering that instance-level candidates are more rational than the redundant 3D object proposals.
		Subsequently, for each candidate, we perform the multi-level contextual inference, i.e., referring from instance attribute perception, instance-to-instance relation perception, and instance-to-background global localization perception, respectively. 
		Eventually, the most relevant candidate is selected and localized by ranking confidence scores, which are obtained by the cooperative holistic visual-language feature matching.
		Experiments confirm that our method outperforms previous state-of-the-arts on ScanRefer online benchmark and Nr3D/Sr3D datasets.
	\end{abstract}
	
	\section{Introduction}
	
	Visual grounding (VG), which aims at localizing the desired objects or areas in an image or a video based on an object-related linguistic query, has achieved great progress in the 2D computer vision community~\cite{kazemzadeh2014referitgame, mao2016generation, wang2018learning, liu2019improving, mogadala2019trends}.
	With the rapid development of 3D sensor and 3D representation, the VG task has gradually merged more informative 3D data.
	Unlike 2D images with regular and well-organized pixels, 3D data mostly comes in the form of point clouds, which is sparse, irregular, and unordered.
	Therefore, previous 2D-based schemes are usually deficient for real 3D scenarios.

	\begin{figure}[t]
		\begin{center}
			\includegraphics[width=0.95\linewidth]{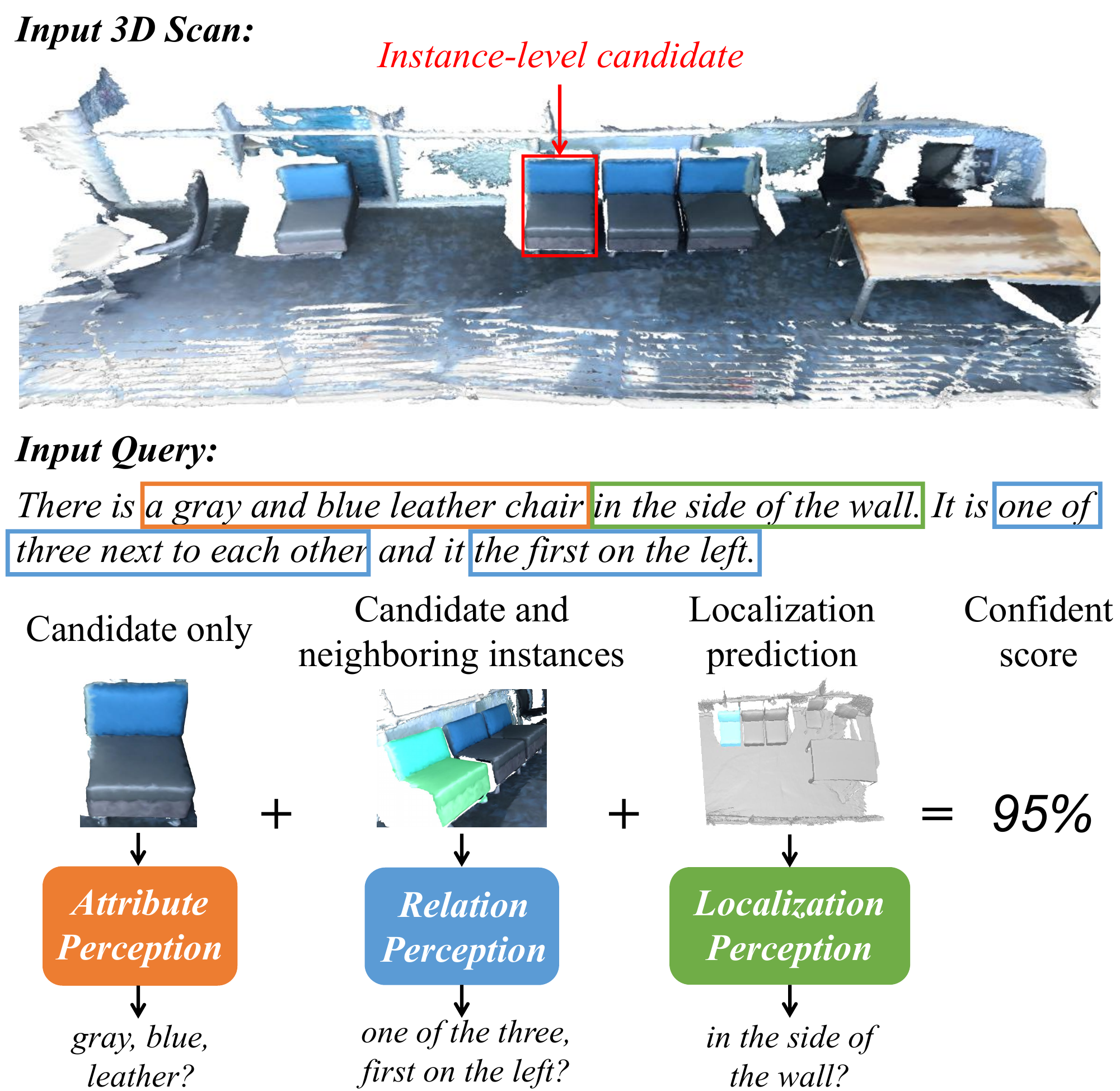}
		\end{center}
		\caption{\textbf{Multi-level contextual referring}. For each instance-level candidate, we match it with linguistic query from attribute, local relation and global localization.
			The attribute, relation and localization descriptions are in orange, blue and green boxes.}
		\label{fig1}
		
	\end{figure}

	Chen~\textit{et al.}~\cite{chen2020scanrefer} is the pioneer for visual grounding on point clouds. They propose the first dataset \textit{ScanRefer} and solve the problem by extending the 2D  grounding-by-detection pipeline to 3D.
	Specifically, it first uses a 3D object detector \cite{qi2019deep} to generate hundreds of proposals. Then the feature of each proposal is merged with a global representation of the linguistic query to predict a matching score. The proposal with the maximal score is considered as the object we are looking for.
	However, it suffers from several issues when transferring the 2D method to 3D VG as follows:
	\textbf{1})  The object proposals in the large 3D scene are usually redundant.
	Compared with the actual instances, the number of proposals is large and the inter-proposal relationship is complex, which inevitably introduces noise and ambiguity.
	%
	\textbf{2})  The appearance and attribute information is not sufficiently captured. 
	Due to the noise and occlusion, the obtained point clouds are usually sparse and incomplete, leading to missing geometric details in object-aware proposals.
	%
	Conventional point cloud-based methods fail to effectively extract the attribute information, \eg, red, gray, and wooden, which might ignore some vital linguistic cues for referring. 
	\textbf{3}) The relations among proposals and the ones between proposals and background are not fully studied.
	
	
	To address the above issues, this paper investigates a novel framework, namely \textbf{InstanceRefer}, to achieve a superior visual grounding on point clouds with \textit{grounding-by-matching} strategy.
	Specifically, via the global panoptic segmentation, our proposed model extracts several instance point clouds from the original scene.
	%
	%
	These instances are subsequently filtered by the predicted category from the natural language descriptions, obtaining the \textit{candidates set}.
	Compared with the object-proposal based candidates~\cite{chen2020scanrefer}, these filtered instance point clouds contain more original geometric and attribute details (\ie, color, texture, \etc) while maintaining a  smaller number.
	We notice that the recent work TGNN~\cite{huang2021text} also employs instance segmentation to reduce the difficulty of referring.
	However, they directly exploit the learned semantic scores from the segmentation backbone as the instance features, which suffer from the lossy geometric and attribute information.
	%
	By comparison, our InstanceRefer applied filtered candidates and their original information for further referring.
	Thus, it can not only reduce the number of the candidates, but also maintain each candidate's original information.
	Besides, to fully comprehend the whole scene, multi-level contextual learning modules are further proposed, \ie, explicitly capturing the context of each candidate from instance attributes, instance-to-instance relationships, and instance-to-background global localization, respectively.
	Eventually, with the well-designed matching module and contrastive strategy, InstanceRefer can efficiently and effectively select and localize the target.
	In consequence, our model outperforms previous methods by a large margin regardless of any settings, \ie, exploiting any segmentation backbone.
	%
	
	In summary, the key contributions of this paper are as follows:
	\textbf{1}) We propose a new framework InstanceRefer for visual grounding on point clouds, which exploits panoptic segmentation and language cues to select the instance point clouds as candidates and re-formulates the task in a grounding-by-matching manner.
	\textbf{2}) Three novel components are proposed to select the most relevant instance candidate from attributes, local relations, and global localization aspects jointly.
	\textbf{3}) Experimental results on \textit{ScanRefer} and Sr3D/Nr3D datasets confirm the superiority of InstanceRefer, which achieves state-of-the-arts on ScanRefer benchmark and Nr3D/Sr3D dataset.

	\section{Related Work}
	
	\noindent\textbf{Visual Grounding on 2D Images.}
	The task of visual grounding on images is to localize a specific area of the image described by a natural language query. 
	Depending on the type of language query, it can be further divided into phrase localization \cite{kazemzadeh2014referitgame, plummer2015flickr30k, wang2018learning} and referring expression comprehension \cite{nagaraja2016modeling, li2017deep, wang2019neighbourhood}. 
	%
	%
	Most approaches conduct localization in the bounding box level and a two-stage manner. 
	The first stage is to generate candidate proposals with either unsupervised methods or a pre-trained object detection network. In the second stage, the best matching proposal is selected according to the language query. 
	Such methods mainly focus on improving the ranking accuracy of the second stage. 
	MAttNet~\cite{yu2018mattnet} proposes a modular attention network to decompose the language query to different components (\ie, subject appearance, location, and relationship to other objects) and process them in different modular networks separately.
	Inspired but different from MAttNet, our work delves specifically into the characteristics of 3D point clouds, and each proposed module differs greatly from those in MAttNet.

	\noindent\textbf{Visual Grounding on 3D Point Clouds.}
	Chen~\textit{et al.}~\cite{chen2020scanrefer} releases the first 3D VG dataset \textit{ScanRefer}, in which the object bounding boxes are referred by their corresponding language queries in an indoor scene.
	ReferIt3D~\cite{achlioptas2020referit3d} also proposes two datasets for 3D VG, Sr3D (labeled by machine) and Nr3D (labeled by human).
	Different from ScanRefer, ReferIt3D assumes that all objects are well-segmented, thus localization is not required.
	Very recently, TGNN~\cite{huang2021text} proposes a similar task called referring 3D instance segmentation, which aims to segment out the target instance.
	It first extracts per-point features and predicts offsets for object clustering.
	Then a  Text-Guided Graph Neural Network is applied to achieve more accurate referring.
	However, TGNN fails to capture the attribute of instances and instance-to-background relations.
	Goyal~\textit{et al.}~\cite{goyal2020rel3d} also presents a dataset named \textit{Rel3D} for only grounding object spatial relations.
	In this paper, we focus on the task of visual grounding on raw point clouds, which is extended from \textit{ScanRefer} and ReferIt3D.

	\begin{figure*}[t]
		\begin{center}
			\includegraphics[width=0.98\linewidth]{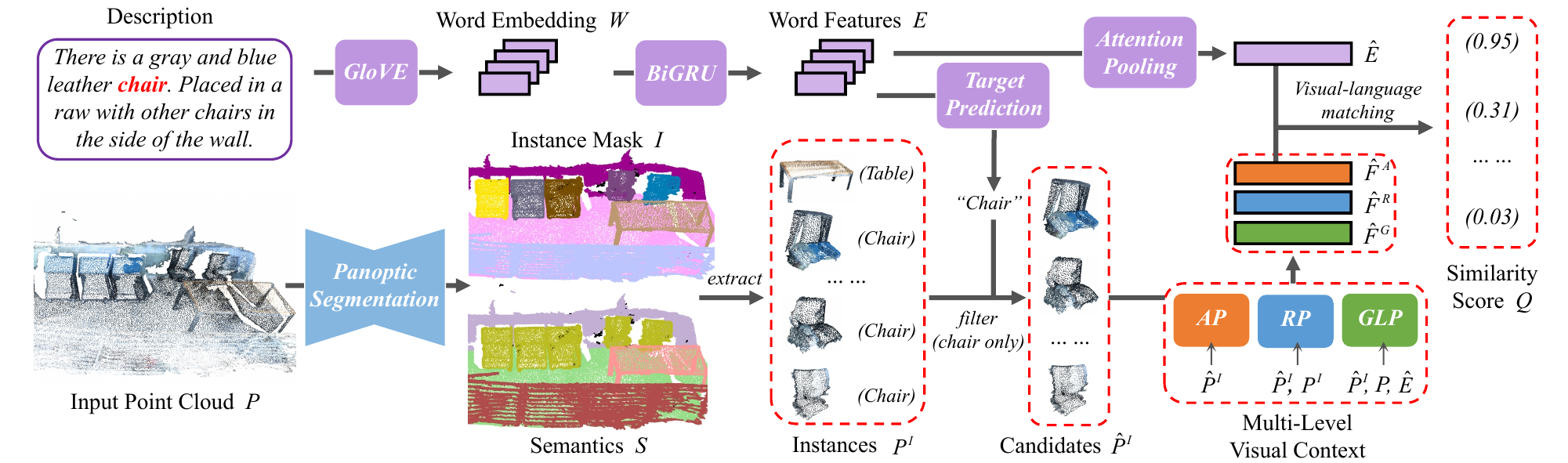}
		\end{center}
		\caption{\textbf{The pipeline of InstanceRefer.} It firstly uses the panoptic segmentation model to extract all the instance point clouds in the large 3D scene. 
			Under the guidance of target prediction from language description, the instances belonging to the target category are filtered out to form the initial candidates $\hat{P}^I$.
			In parallel, the summarized language encoding $\hat{E}$ is achieved through attention pooling.
			Subsequently, a visual-language matching module outputs the similarity score $Q$ through comparing multi-level visual perceptions (\ie, $\hat{F}^{A}$, $\hat{F}^{R}$, and $\hat{F}^{G}$) against $\hat{E}$.
			Eventually, 3D bonding-box of the instance with the highest score is regarded as the final grounding result.}
		\label{fig2}
	\end{figure*}
	
	\noindent\textbf{3D Representation Learning on Point Clouds.} 
	Unlike 2D images with regular grids, point clouds are irregular and often sparsely-scattered.
	%
	%
	Recently, point-based models leverage the permutation invariant nature of raw point cloud for enhanced 3D processing~\cite{qi2017pointnet,qi2017pointnet++}.
	Specifically, most point-based models first sample sub-points from the initial point clouds, then apply a feature aggregation function on each sub-point in a local point cloud cluster after the grouping.
	The representatives of the point-based methods are graph-based learning~\cite{wang2019dynamic,zhiheng2019pyramnet,landrieu2018large,landrieu2019point} and convolution-like operations~\cite{zhao2019pointweb,Thomas_2019_ICCV,PointConv,hu2019randla,yan2020pointasnl}.
	With the development of the representation learning on point clouds, various downstream tasks related to visual grounding on point clouds have been explored rapidly, \eg, 3D object detection~\cite{qi2019deep} and instance segmentation~\cite{jiang2020pointgroup}.  

	\section{Method}
	
	
	\label{sec:method}
	InstanceRefer is a novel framework for point cloud VG, which conducts the multi-level contextual referring to select the most relevant instance-level object.
	As shown in Figure~\ref{fig2}, by exploiting point cloud panoptic segmentation, InstanceRefer first extracts instances with their semantic labels from the whole scene (Sec.~\ref{sec2}).
	Following that, the sentences are embedded into a  high-dimensional feature space, and a text classification is conducted as the linguistic guidance (Sec.~\ref{sec3}).
	Finally, after filtering out the candidates from all instances, the three-level progressive referring modules, \ie, attribute perception (AP) module, relation perception (RP) module, and global localization perception (GLP) module (Sec.~\ref{sec4}), are employed to select the optimal candidate.
	
	\subsection{Instance Set Generation}
	\label{sec2}
	Unlike ScanRefer~\cite{chen2020scanrefer} that selects the candidates from all object proposals, our framework first extracts all foreground instances from the input point cloud to generate a set of instances.
	Then, we re-formulate the 3D visual grounding problem as an instance-matching problem.
	
	For this purpose, panoptic segmentation~\cite{kirillov2019panoptic} is adopted in our model, which aims to tackle the semantic segmentation and the instance segmentation jointly.
	Taking a point cloud $P \in \mathbb{R}^{N\times 3}$ and its features $F \in \mathbb{R}^{N\times D}$ as input, InstanceRefer returns two prediction sets, semantics $S \in \mathbb{R}^{N\times 1}$ and instance masks $I \in \mathbb{R}^{N\times 1}$, recording the semantic class and the instance index of each point, respectively.
	Through the instance masks, InstanceRefer extracts instance point clouds from the original scene point clouds. 
	All instance point clouds in the foreground are represented as $\boldsymbol P^{I} = \{P^{I}_i\}_{i=0}^M$, where $P^{I}_i$ means the points of the $i$-th instance within the total $M$ instances.
	Similarly, features and semantics of all instances are denoted as $\boldsymbol F^{I}$ and $\boldsymbol S^{I}$, respectively.

	\subsection{Description Encoding}
	\label{sec3}
	Each token of the language description is first mapped into the 300d-vector via the pre-trained GloVE word embedding~\cite{pennington2014glove}. Then the whole sequence is fed into Bidirectional GRU layers~\cite{chung2014empirical} to extract the contextual word features $E=\{e_i\} \in \mathbb{R}^{N_w\times D}$, where $N_w$ is the query length and $D$ is the feature dimension. 
	The final language encoding is achieved through attention pooling.
	%
	In practice, the attention pooling updates each word feature and aggregate them to a global representation by
	\begin{align}
	\hat{e}_{i} &= \texttt{AvgPool}(\{\texttt{Rel}(e_{i} ,e_{j}) \odot e_{j}, ~~ \forall~e_{j} \in E\}),\\
	\hat{E} &= \texttt{MaxPool}(\{\hat{e}_{i}\}_{i=1}^{N_w}),
	\label{eq1}
	\end{align}
	where the aggregation function $\texttt{AvgPool}(\cdot)$ and $\texttt{MaxPool}(\cdot)$ are set as average pooling and max pooling and a pairwise relationship function $\texttt{Rel}(\cdot)$ is the normalized dot-product similarity between features of two tokens, and the sign $\odot$ represents the element-wise multiplication.
	%
	The feature of each token $e_i \in E$ is first updated to $\hat{e}_i$ via the aggregation of all token features weighted by relation.
	Then, the global representation of the query is obtained, \ie, $\hat{E} \in \mathbb{R}^{1\times D}$.
	%
	%
	Furthermore, by appending an additional GRU layer and linear layer, InstanceRefer predicts the target category of the query by language features. 
	This output aids the model in subsequently filtering out the candidates from all instances.
	
	\subsection{Multi-Level Visual Context}
	\label{sec4}
	
\begin{figure}[t]
		\begin{center}
			\includegraphics[width=0.95\linewidth]{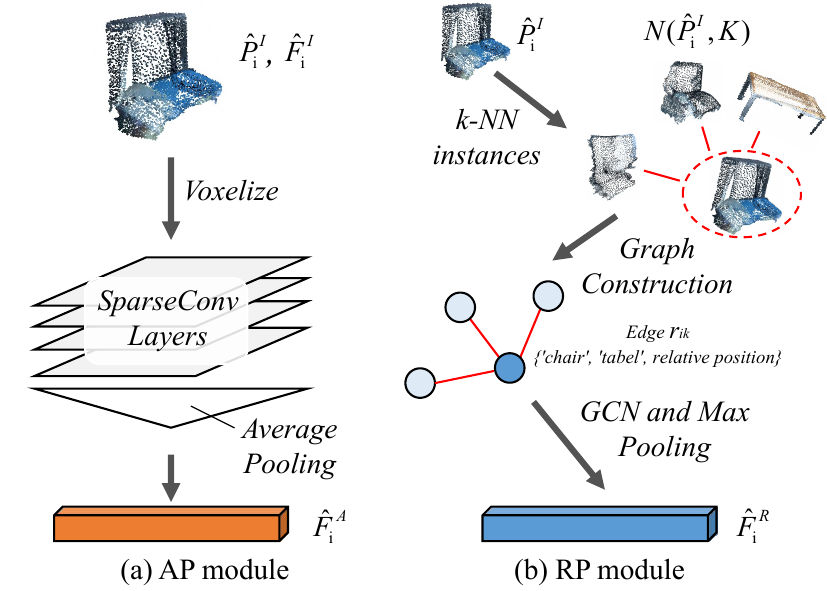}
		\end{center}
		\caption{\textbf{The inner structure of AP and RP modules.} 
			Part (a) illustrates the attribute perception (AP) module, which uses a four-layer SparseConv to extract the global features of each candidate.
			In part (b), the relation perception (RP) module aggregates the information of the candidate with its neighboring instances.
		}
		\label{fig3}
	\end{figure}

\begin{figure}[t]
		\begin{center}
			\includegraphics[width=0.9\linewidth]{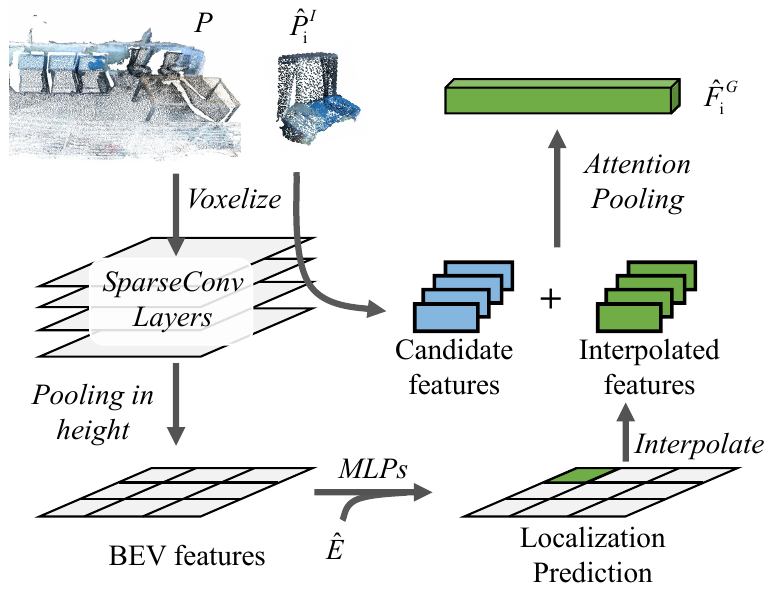}
		\end{center}
		\caption{\textbf{Illustration of GLP module.}
			GLP module firstly predicts the localization of the target in a $3\times 3$ bird-eyes-view (BEV) plane.
			It then uses the interpolated features and candidate features to generate the aggregated feature $\hat{F}^G_i$.}
		\label{fig4}
	\end{figure}

	Before feeding the instances into the following modules, InstanceRefer first uses the predicted target category from the language encoder to filter \textit{candidates}.
	For example, as shown in Figure~\ref{fig2}, regarding all instances extracted from the original point cloud, we only keep the remaining instances belonging to the target category `chair'.
	Subsequently, the corresponding point clouds and features of candidates $\hat{\boldsymbol P}^{I}$ and $\hat{\boldsymbol F}^{I}$ are attained.
	Note that the target classification accuracy of language query is over 97\%.
	Hence, the exploited filtering operation will not introduce obvious noise for further referring, while it can greatly boost the grounding performance for the unique instance candidate scenario.
	Then, the filtered instances are compared with the following multi-level visual context modules.\\
	%
	\noindent\textbf{AP Module.} 
	Considering there are many adjectives in a sentence (\eg, ``\textit{a long bookshelf}" for scale,  ``\textit{a brown chair}" for color, ``\textit{a square table}" for shape,~\etc), the attribute perception (AP) module is designed to explicitly capture such information from attribute phrases.
	Concretely, the AP module takes the information of the $i$-th candidate, \ie, point cloud $\hat{P}^{I}_i$ and its attribute features $\hat{F}^{I}_i$ as input and generates a global representation vector $F^{A}_i$ of the candidate.
	As shown in Figure~\ref{fig3} (a), our model constructs a four-layer Sparse Convolution (SparseConv)~\cite{graham20183d} as the feature extractor.
	The extractor firstly voxelizes the point cloud into 3D voxels and then conducts the convolution operation only on non-empty voxels in a more efficient way.
	%
	%
	Subsequently, through an average pooling, the feature representation $\hat{F}^A \in \mathbb{R}^{1\times D}$ is obtained.
	\\[.2cm]
	\noindent\textbf{RP Module.}
	Since there are many descriptions about the relations between different instances, \eg, ``\textit{The desk is between a black chair and bed}", using only the attribute-related manner fails to capture such information. 
	Therefore, a relation perception (RP) module is proposed for the relation encoding between the candidate and its surrounding instances.
	Figure~\ref{fig3} (b) illustrates the design of the RP module.
	Given the $i$-th candidate point cloud $\hat{P}^{I}_i$, the RP module first searches $K$ instance-level neighborhoods that have the closest Euclidean distance to the center of instance $\hat{P}^{I}_i$.
	Following that, a graph-based aggregation is employed to fuse the features of local neighborhoods.
	To define the edge-relation ${r}_{ij}$ between the $i$-th candidate and the $k$-th neighboring instance, the RP module adopts the DGCNN~\cite{wang2019dynamic}:
	\begin{equation}
	{r}_{ik} = \texttt{MLP}([\mathcal{C}(\hat{P}^{I}_i)-\mathcal{C}({P}^{I}_k); S^I_i; S^I_k]),
	\label{eq6}
	\end{equation}                    
	where $\mathcal{C}(\cdot)$ is to choose center coordinates of instances, $S_i^I$ and $S_k^I$ represent the semantic instance masks for the $i$-th and $k$-th instances, respectively. $\texttt{MLP}(\cdot)$ denotes the multi-layer perceptrons.
	The sign $[\cdot;\cdot]$ means the channel-wise concatenation.
	%
	%
	Through Eq.~\eqref{eq6}, the RP module considers not only the relative positions but also the semantic relationships between the candidate and its neighbors.
	Eventually, the enhanced feature $\hat{F}^R_i \in \mathbb{R}^{1\times D}$ for the $i$-th candidate can be obtained by
	\begin{align}
	h_{ik} &= \texttt{MLP}([{P}^{I}_k; S^I_k]), ~ \forall~{P}^{I}_k \in \mathcal{N}(\hat{P}^{I}_i, K),\\ 
	\hat{F}^R_i &= \texttt{MaxPool}(\{{r}_{ik} \odot h_{ik}\}_{k=1}^K),
	\label{eq7}
	\end{align}
	where $\mathcal{N}(\hat{P}^{I}_i, K)$ denotes the $K$ nearest neighboring instances for $\hat{P}^{I}_i$, and $h_{ik}$ is the unified representation of the coordinate and the semantic instance mask for the $k$-th instance. 
	Note that the above $\texttt{MLP}(\cdot)$ have the same output dimensions with that in Eq.~\eqref{eq6}.
	\\[.2cm]
	\noindent\textbf{GLP Module.}
	The global localization perception (GLP) module aims to supplement the background information neglected by the two aforementioned modules.
	There are other descriptions about global localization information, \eg, ``\textit{in the corner}" and ``\textit{next to the wall}", but such information cannot be included in the AP and RP modules.
	As shown in Figure~\ref{fig4}, the GLP module takes the entire point cloud as the input.
	By employing another SparseConv encoder, the module first extracts the point-wise features of the entire scene.
	An average pooling in the height axis is then performed to generate the bird-eyes-view (BEV) features.
	Note that each input point cloud is divided into $3\times 3$ areas in the BEV plane.
	By repeatedly concatenating the language features $\hat{E}$ and then flowing through the MLPs, the GLP module predicts the object candidate's location in one of the nine areas.
	Moreover, the probability of each area is interpolated into point clouds of the $i$-th candidate $\hat{P}^I_i$ by
	\begin{align}
	\hat{f}_{i,k} &= \sum_{j=1}^{n}||\hat{p}^{I}_{i,k}-a^{\text{c}}_j||^{-1}_2 \cdot a^{\text{p}}_j,
	\label{eqbev}
	\end{align} 
	where $\hat{p}^{I}_{i,k} \in \hat{P}^I_i$ and $\hat{f}_{i,k}$ are the coordinates and interpolated features of the $k$-th point in the $i$-th candidate.
	$n = 9$ represents the total number of areas, $a^{\text{p}}_j$ and $a^{\text{c}}_j$ are the probabilities of localization and center coordinates of $j$-th area in the total nine areas, respectively.
	Finally, after concatenating interpolated features with candidate features, another MLPs with max-pooling aggregates the features of the $i$-th candidate to a global representation $\hat{F}_i^G$.
	
	\subsection{Visual-Language Matching}
	With multi-level visual features of the candidates (\ie, $\hat{\boldsymbol F}^A$, $\hat{\boldsymbol F}^R$, $\hat{\boldsymbol F}^G$) and language features $\hat{E}$, we perform a matching operation to obtain the confidence score for each instance.
	Considering the obtained multiple features, a simple scheme of pinpointing the referred target is to find the most relevant visual features to the linguistic ones through their similarities.
	However, this approach ignores the varying proportions for three perception modules.
	To tackle this issue, we utilize the modular co-attention from MCAN~\cite{yu2019mcan} to perform adaptive visual-language matching. 
	For the $i$-th instance, we concatenate three visual features to the merged features $\hat{F}_i\in \mathbb{R}^{1\times (D \times 3)}$. Then, we further employ three co-attention layers to aggregate the language features to update the instance feature. Finally, the \textit{sigmoid} activation function is exploited to output the instance score.

	\subsection{Contrastive Objective}
	For the objective function, we adopt a \textit{contrastive manner} to train our network. 
	Here, we define an instance as a positive example for a query if its IoU with the GT object bounding box exceeds the threshold $\Gamma$, otherwise a negative example.  
	If there is no positive example for a query, we do not compute its loss.
	Intuitively, the matching score of positive pairs should be higher than negative pairs. 
	Thus we derive our matching loss from \cite{sohn2016improved} with the consideration of multiple positive examples as follows:
	\begin{equation}
	L_{\text{mat}} = - \text{log} \frac{\sum_{i=1}^{L} \text{exp}(Q_i^+)}{\sum_{i=1}^{L} \text{exp}(Q_i^+) + \sum_{i=L+1}^{M} \text{exp}(Q_i^-)},
	\label{eq9}
	\end{equation}
	where $Q^+$ and $Q^-$ denote the scores of positive and negative pairs, $L$ and $M$ are the numbers of positive candidates and total candidates in a scene, respectively.
	All candidates are introduced to the optimization process.
	
	The language classification loss $L_{\text{cls}}$ and the BEV localization loss $L_{\text{bev}}$ are also included for the joint target categorization and localization prediction. 
	The final loss is a weighted sum of matching loss, including the object classification loss on the linguistic queries and the localization loss,
	$L = L_{\text{mat}} + \lambda_1 L_{\text{cls}} + \lambda_2 L_{\text{bev}}$, where $\lambda_1 = \lambda_2 = 0.1$ are the weights to adjust the ratios of each loss, respectively. The IoU threshold $\Gamma$ is set as 0.3.
	
	\begin{table*}[htbp]
		\centering 
		\vspace{.1cm}
		\caption{
			Comparison of localization results. TGNN replaces the original GRU layers with pre-trained BERT to extract language features. 
			Our method follows TGNN's strategy of only taking coordinates (Geo) and color information (RGB) as input, while results of ScanRefer on benchmark are obtained by using additional normals (Nor) and multi-view features from a pre-trained 2D feature extractor.
			Scores for the test set are obtained from the online evaluation. Only the published methods are compared. Accessed on March 18, 2021.
		}
		\small
		\begin{tabular}{lc|cccccc}
			\toprule  
			& & \multicolumn{2}{c}{Unique} & \multicolumn{2}{c}{Multiple} & \multicolumn{2}{c}{Overall}\\
			Method & Input & Acc@0.25 & Acc@0.5 & Acc@0.25 & Acc@0.5 & Acc@0.25 & Acc@0.5 \\
			\midrule  
			\multicolumn{8}{c}{Validation results} \\
			\midrule  
			SCRC \cite{hu2016natural} & RGB image & 24.03 & 9.22 & 17.77 & 5.97 & 18.70 & 6.45 \\
			One-stage \cite{yang2019fast} & RGB image & 29.32 & 22.82 & 18.72 & 6.49 & 20.38 & 9.04 \\
			\midrule
			ScanRefer \cite{chen2020scanrefer} & Geo + RGB & 65.00 & 43.31 & 30.63 & 19.75 & 37.30 & 24.32 \\
			TGNN \cite{huang2021text} & Geo + RGB & 64.50 & 53.01 & 27.01 & 21.88 & 34.29 & 27.92 \\
			TGNN\cite{huang2021text}+BERT \cite{devlin2018bert}  & Geo + RGB & 68.61 & 56.80 & 29.84 & 23.18 & 37.37 & 29.70 \\
			IntanceRefer (Ours) & Geo + RGB & \textbf{77.45} & \textbf{66.83} & \textbf{31.27} & \textbf{24.77} & \textbf{40.23} & \textbf{32.93} \\
			
			\midrule  
			\multicolumn{8}{c}{Test results (ScanRefer benchmark)} \\
			\midrule  
			ScanRefer \cite{chen2020scanrefer}& Geo+Nor+Multiview & 68.59 & 43.53 & 34.88 & 20.97 & 42.44 & 26.03 \\
			\midrule
			TGNN \cite{huang2021text} & Geo + RGB & 62.40 & 53.30 & 28.20 & 21.30 & 35.90 & 28.50 \\
			TGNN \cite{huang2021text}+BERT \cite{devlin2018bert} & Geo + RGB & 68.34 & 58.94 & 33.12 & 25.26 & 41.02 & 32.81 \\
			IntanceRefer (Ours) & Geo + RGB & \textbf{77.82} & \textbf{66.69} & \textbf{34.57} & \textbf{26.88} & \textbf{44.27} & \textbf{35.80} \\

			\bottomrule 
		\end{tabular}
		\label{tab1}

	\end{table*}

	\section{Experiments}
	
	In this section, we present the experimental procedures and analysis in detail to demonstrate the effectiveness of our InstanceRefer in 3D visual grounding.
	
	\subsection{Implementation}
	In our experiment, we adopt the official pre-trained PointGroup~\cite{jiang2020pointgroup} as the backbone to perform the panoptic segmentation.
	For language encoding, we employ the same GloVE and BiGRU used in ScanRefer~\cite{chen2020scanrefer} to generate the word features in channel $D=256$.
	The output of self-attention preserves the identical 256 channels.
	The AP module consists of four 3D sparse convolution blocks, each of which has two 3D sparse convolutions inside. 
	Going deeper, we gradually increase the number of channels (\ie, 32, 64, 128, 256).
	The GLP module applies the same structure of the sparse convolution block but with fewer blocks (\ie, 3 blocks with channel 32, 128, 256).
	In the RP module, the kNN instance number $K$ is 8, and the channel numbers of two MLPs are (256, 256) and (256), respectively.

	We train the network for 30 epochs by using the Adam optimizer with a batch size of 32. 
	The learning rate of the network is initialized as 0.0005 with the decay as 0.9 for every 10 epochs. 
	All experiments are implemented on PyTorch and a single NVIDIA 1080Ti GPU. 
	We will release our code and pre-trained model for future research.

	\subsection{Dataset and Metrics}
	\noindent\textbf{ScanRefer.}  The \textit{ScanRefer} dataset is a newly proposed 3D scene visual grounding dataset, to the best of our knowledge, which consists of 51,538 descriptions for ScanNet scenes~\cite{scannet}. 
	The dataset is split into 36,655 samples for training, 9,508 samples for validation, and 5,410 samples for testing, respectively.
	
	For the evaluation metrics, it calculates the 3D intersection over union (IoU) between the predicted bounding box and ground truth.
	The Acc@$m$IoU is adopted as the evaluation metric, where $m \in \{0.25, 0.5\}$.
	Accuracy is reported in ``unique" and ``multiple" categories, respectively.
	If only a single object of its class exists in the scene, we regard it as ``unique", otherwise ``multiple".
	Moreover, to fully evaluate our model, we conduct a fair comparison on both the validation set and test set. 
	\textit{ScanRefer} benchmark\footnote{\url{http://kaldir.vc.in.tum.de/scanrefer_benchmark}} conducts online testing and every method is allowed to submit results for \textit{only twice}.\\[.1cm]
	\noindent\textbf{Nr3D and Sr3D.}  
	\textit{ReferIt3D}~\cite{achlioptas2020referit3d} dataset uses the same train/valid split with \textit{ScanRefer} on ScanNet but exploits manually extracted instances as input, \ie, object masks for each scene are provided, and aims to choose the solely referred object.
	Specifically, it contains two datasets, where Sr3D (Spatial Reference in 3D) has 83.5K synthetic expressions generated by templates and Nr3D (Natural Reference in 3D) consists of 41.5K human expressions collected in a similar manner as ReferItGame~\cite{kazemzadeh2014referitgame}. 
	Since \textit{ReferIt3D} directly uses point clouds of instances as input, it can be seen as the instance-matching stage of our InstanceRefer without interaction with the environment, \ie, wall and floor.
	We empirically validate AP and RP modules on \textit{ReferIt3D} to verify the effectiveness of our proposed modules.
	We use the same evaluation strategies and metrics with their paper. 
	
	\begin{table*}[t]
		\centering 
		\footnotesize
		\caption{Comparison of referring object identification on Nr3D and Sr3D datasets. Here `easy' and `hard' determined by whether there are more than two instances of the same object class in the scene. `view-dependent' and `view-independent' determined by whether the referring expression depending on camera view.}
		\vspace{.1cm}
		\begin{tabular}{llccccc}
			\toprule  
			Dataset & Method & Easy & Hard & View-dep. & View-indep. & Overall \\
			\midrule  
			\multirow{3}{0.6cm}{Nr3D}& ReferIt3DNet~\cite{achlioptas2020referit3d}  & \SI{43.6 \pm 0.8}{\percent} & \SI{27.9 \pm 0.7}{\percent} & \SI{32.5 \pm 0.7}{\percent} & \SI{37.1 \pm 0.8}{\percent} & \SI{35.6 \pm 0.7}{\percent} \\
			& TGNN \cite{huang2021text}  & \SI{44.2 \pm 0.4}{\percent} & \SI{30.6 \pm 0.2}{\percent} & \textbf{\SI{35.8 \pm 0.2}{\percent}} & \SI{38.0 \pm 0.3}{\percent} & \SI{37.3 \pm 0.3}{\percent} \\
			& IntanceRefer (Ours)  & \textbf{\SI{46.0 \pm 0.5}{\percent}} & \textbf{\SI{31.8 \pm 0.4}{\percent}} & \SI{34.5 \pm 0.6}{\percent} & \textbf{\SI{41.9 \pm 0.4}{\percent}} & \textbf{\SI{38.8 \pm 0.4}{\percent}} \\
			\midrule  
			\multirow{3}{0.6cm}{Sr3D}& ReferIt3DNet~\cite{achlioptas2020referit3d}  & \SI{44.7 \pm 0.1}{\percent} & \SI{31.5 \pm 0.4}{\percent} & \SI{39.2 \pm 1.0}{\percent} & \SI{40.8 \pm 0.1}{\percent} & \SI{40.8 \pm 0.2}{\percent} \\
			& TGNN \cite{huang2021text}  & \SI{48.5 \pm 0.2}{\percent} & \SI{36.9 \pm 0.5}{\percent} & \textbf{\SI{45.8 \pm 1.1}{\percent}} & \SI{45.0 \pm 0.2}{\percent} & \SI{45.0 \pm 0.2}{\percent} \\
			& IntanceRefer (Ours) & \textbf{\SI{51.1 \pm 0.2}{\percent}} & \textbf{\SI{40.5 \pm 0.3}{\percent}} & \SI{45.4 \pm 0.9}{\percent} & \textbf{\SI{48.1 \pm 0.3}{\percent}} & \textbf{\SI{48.0 \pm 0.3}{\percent}} \\
			\bottomrule 
		\end{tabular}
		\label{tab1_1}
	\end{table*}
	
	\begin{figure*}[t]
		\begin{center}
			\includegraphics[width=0.95\linewidth]{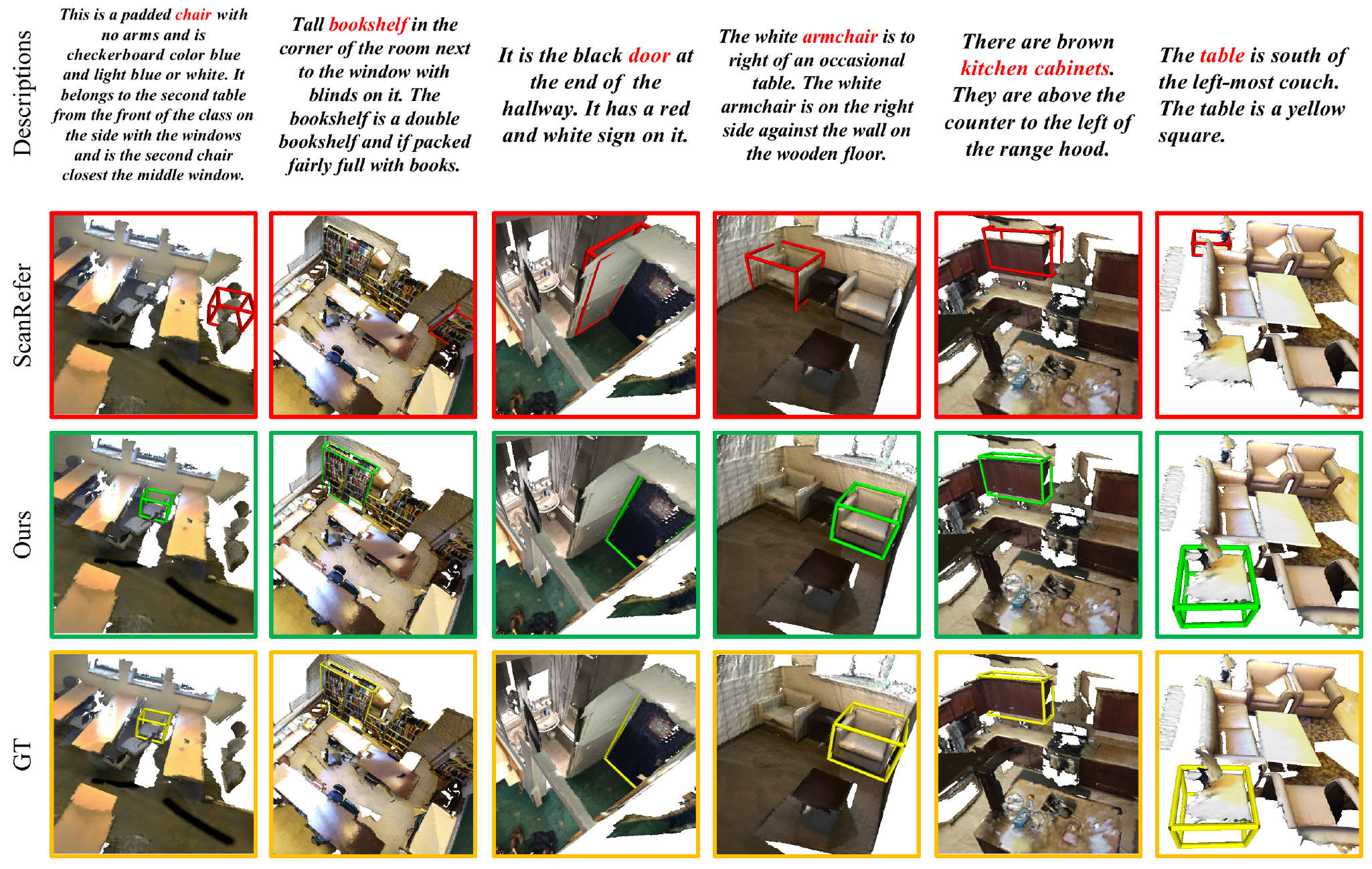}
		\end{center}
		\caption{Qualitative results from ScanRefer and our InstanceRefer. Predicted boxes are marked {\color{green}{green}} if they have an IoU score higher than 0.5, otherwise they are marked {\color{red}{red}}. The ground truth boxes are displayed in {\color{yellow}{yellow}}.}
		\label{fig5}
	\end{figure*}
	
	\subsection{Quantitative Comparisons}
	
	We first compare IntanceRefer with the state-of-the-art methods on the \textit{ScanRefer} dataset, where the results are displayed in Table~\ref{tab1}.
	Among these methods, SCRC~\cite{hu2016natural} and One-stage~\cite{yang2019fast} are image-based methods with the RGB images as input.
	Specifically, they select the 2D bounding box with the highest confidence score and project it to the 3D space using the depth map of that frame.
	ScanRefer~\cite{chen2020scanrefer} and TGNN~\cite{huang2021text} are point cloud based methods that take coordinates and other features of point clouds as input.
	In this paper, we follow the input modality of TGNN~\cite{huang2021text}, which only exploits geometric coordinates (XYZ) and color information (RGB) as input.
	
	As shown in Table~\ref{tab1}, our model gains the highest scores on both validation set and online benchmark.
	Note that the image-based methods (\ie, SCRC and One-stage) fail to achieve satisfactory results since they are limited by the view of a single frame.
	Though TGNN also applies PointGroup to perform instance segmentation, our method outperforms it by large margins, especially in the ``unique" case, which mainly benefits from the rational strategy of filtering candidates.
	Furthermore, the improvements of our model are apparent, reporting 11.8\% in ``unique" and 6.7\% ``multiple" in Acc@0.5 when employing both the GloVE and GRU as the language encoder.
	Our improvements originate from the well-designed pipeline and three novel models, whereas the improvements of TGNN are largely based on the pre-trained BERT embeddings.
	More importantly, even if ScanRefer utilizes additional multiview features from the pre-trained ENet~\cite{paszke2016enet} on the benchmark, the overall result of ours still achieves $\sim$10\% improvement on Acc@0.5.
	Also, since we employ a pre-trained panoptic segmentation model and stores the segmentation results offline, our method has a much shorter training time and lower memory consumption than others.
	
	The results in Table~\ref{tab1_1} illustrate the instance-matching accuracy on the Nr3D and Sr3D datasets.
	Our proposed {{InstanceRefer}} achieves top-ranked results on both the Nr3D and Sr3D datasets.
	The experiments prove that our proposed perception modules are effective components for accurate grounding on the scenes of point clouds and boost the pure instance-matching performances significantly.
	As a result, our {{InstanceRefer}} manifests a stronger capacity than ScanRefer and TGNN on the 3D point cloud VG task.
	
	Visualization results produced by ScanRefer and by our method are displayed in Figure~\ref{fig5}.
	Predicted boxes are marked green if they have an IoU score higher than 0.5, otherwise, they are marked red.
	Failure cases of ScanRefer illustrate that its architecture cannot distinguish ambiguous objects according to their spatial relations.
	On the contrary, InstanceRefer can accurately localize the described objects even in the complex scenarios with the long textual descriptions, \eg, the results in the first and second columns show our accurate instance selection and the fifth columns illustrates our method can generate bounding box more finely. 
	%
	
	\begin{table}[t]
		
		\caption{The ablated results for different network architecture on \textit{ScanRefer} validation set, where Acc@0.5 is used as metrics. Here MAT. means matching module.}
		\vspace{.1cm}
		\centering 
		\small
		\begin{tabular}{cccc|ccc}
			\toprule  
			
			AP & RP & GLP & MAT.& {Unique} & {Multiple} & {Overall}\\\hline
			\cmark&  &&& 66.43  & 20.32  & 29.46 \\
			& \cmark &&& 62.66  & 19.67  & 28.01 \\
			&  &\cmark&& 62.85  & 16.21  & 25.29 \\\hline
			\cmark& \cmark &&& 66.59  & 21.56  & 30.49 \\
			\cmark& \cmark &\cmark&& 66.80  & 22.18  & 31.04 \\
			\cmark& \cmark &\cmark&\cmark& \textbf{66.83}  & \textbf{24.77}  & \textbf{32.93} \\
			
			\bottomrule 
			\label{ablation}
		\end{tabular}
		\label{tab2}
		\vspace{-.3cm}
	\end{table}

	\subsection{Comprehensive Analysis}
	\noindent\textbf{Ablation Studies.}
	Table~\ref{tab2} presents the effectiveness of different modules.
	On the one hand, if only a single perception module is employed, the AP module can achieve the best results.
	On the other hand, when the additional RP and GLP modules are utilized, a remarkable improvement can be seen over the AP module's result.
	Specifically, the gain of the RP module is slightly larger than that of the GLP module.
	The reason is that the descriptions of the correlation among instances in the scene are more commonly compared with the localization. 
	Furthermore, the visual-language matching method we apply is better than simple ranking by cosine similarity. \\[.15cm]
	%
	%
	\noindent\textbf{Instance-matching with Same Backbone.}
	To further illustrate the effectiveness of the proposed modules, we compare instance-matching results when using instances extracted by the same panoptic segmentation backbone or ground truth instances. 
	For the ScanRefer \cite{chen2020scanrefer}, we use PointNet++~\cite{qi2017pointnet++} to extract features of each instance and replace its proposal features. Also, we evaluate ScanRefer using points in ground truths bounding boxes as input.
	For the ReferIt3DNet \cite{achlioptas2020referit3d}, since its original framework is applied on manually segmented instances, we use its original model directly.
	Both of them are trained using the same strategies of our model for the fair comparison.
	Since TGNN \cite{huang2021text} originally used PointGroup to conduct instance segmentation, we will not discuss it in this section.
	
	The experimental results are summarized in Table~\ref{backbone}.
	From the upper part of Table~\ref{backbone}, we can find that panoptic segmentation by PointGroup~\cite{jiang2020pointgroup} can boost the performances of ScanRefer, especially on the ``unique'' case.
	Besides, ReferIt3DNet achieves similar performances with ScanRefer when using the extracted instances by PointGroup. 
	Note that InstanceRefer still surpasses them by a significant improvement on Acc@0.5, achieving {\textbf{6.6\%}} in for ``unique'' and {\textbf{3.1\%}} for ``multiple''.
	It confirms that our gains not only come from the filtering strategy based on panoptic segmentation but also the well-designed multi-level perception modules. 
	Besides, the improvement using ground-truth target is limited since the language classification is high enough (over 97\%).
	%
	%
	From the bottom part of Table~\ref{backbone}, we compare different methods using ground truth instances as input.
	Experimental results reveal that using GT instance point clouds is better than using points in GT bounding boxes, which partly owes to the interference of occlusion in 3D bounding boxes.
	Also, InstanceRefer achieves state-of-the-art performances over ReferIt3DNet and ScanRefer, which credits to the instance-matching strategy.
	In summary, our proposed framework can effectively boost the performances.

	\begin{table}[t]
		
		\caption{The instance-matching results with same panoptic segmentation backbone on \textit{ScanRefer} validation set, where Acc@0.5 is used as metrics. $^\star$ denotes using GT language classification.}
		\small
		\vspace{.1cm}
		\centering 
		
		\begin{tabular}{lccc}
			\toprule  
			Backbone \& Method & {Unique} & {Multiple} & {Overall}\\\hline
			ScanRefer & 43.31  & 19.75  & 24.32 \\
			PG \cite{jiang2020pointgroup} + ScanRefer& 60.05  & 21.61  & 29.07 \\
			PG \cite{jiang2020pointgroup} + ReferIt3DNet& 60.22  & 21.41  & 28.94 \\
			InstanceRefer& {66.83}  & {24.77}  & {32.93} \\
			InstanceRefer$^\star$& \textbf{68.78} & \textbf{24.82} & \textbf{33.35} \\\hline
			GT Box + ScanRefer & 73.55 & 32.00 & 40.06 \\
			GT Inst + ScanRefer & 79.35 & 36.08 & 44.48 \\
			GT Inst + ReferIt3DNet& 79.04 & 37.19 & 45.38 \\
			GT Inst + InstanceRefer& \textbf{90.24} & \textbf{39.32}  & \textbf{49.20} \\
			
			\bottomrule 
			\label{backbone}
		\end{tabular}
		\vspace{-.3cm}
	\end{table}

	\section{Conclusion}
	In this paper, we proposed a novel framework, named InstanceRefer, for 3D visual grounding.
	Our model performs more accurate localization via unifying instance attribute, relation, and localization perceptions.
	Specifically, InstanceRefer innovatively predicts the target category from linguistic queries and filters out a small number of candidates by panoptic segmentation.
	Moreover, we propose the concept of cooperative holistic scene-language understanding for each candidate, \ie, multi-level contextual referring to instance attribute, instance-to-instance relation, and instance-to-background global localization. 
	Experimental results demonstrate that InstanceRefer outperforms the previous methods by a large margin.
	We believe our work formulates a new strategy for 3D visual grounding.

	\section*{Acknowledgment}
	
	{\noindent This work was supported in part by NSFC-Youth  61902335,  by Key Area R\&D Program of Guangdong Province with grant No.2018B030338001, by the National Key R\&D Program of China with grant No.2018YFB1800800,  by Shenzhen Outstanding Talents Training Fund, by Guangdong Research  Project No.2017ZT07X152, by Guangdong Regional Joint Fund-Key Projects 2019B1515120039,  by the  NSFC 61931024\&81922046, by helixon biotechnology company Fund and CCF-Tencent Open Fund.}
	
	{
		\bibliographystyle{ieee_fullname}
		\bibliography{egbib}
	}
	
\end{document}